\def\year{2020}\relax
\documentclass[letterpaper]{article} 
\usepackage{aaai20}  
\usepackage{times}  
\usepackage{helvet} 
\usepackage{courier}  
\usepackage[hyphens]{url}  
\usepackage{graphicx} 
\usepackage{graphicx}  
\usepackage{epsfig}
\usepackage{amsmath}
\DeclareMathOperator*{\argmax}{argmax}
\usepackage{amssymb}
\usepackage{multirow}
\usepackage{booktabs}
\usepackage{subcaption}
\usepackage[export]{adjustbox}
\usepackage{inputenc}
\usepackage{algorithm}
\usepackage{algorithmic}

\usepackage{enumitem}
\setlist[itemize]{leftmargin=*}
\usepackage[dvipsnames]{xcolor}

\newcommand{\citet}[1]{\citeauthor{#1} \shortcite{#1}}

\usepackage[pagebackref=false,breaklinks=false,bookmarks=false]{hyperref}

\title{Character Matters: Video Story Understanding with Character-Aware Relations}

\author{Anonymous Author(s)}

\author{Shijie Geng\textsuperscript{\rm 1},  \Large \textbf{Ji Zhang\textsuperscript{\rm 1},} \Large \textbf{Zuohui Fu\textsuperscript{\rm 1},} \Large \textbf{Peng Gao\textsuperscript{\rm 2}, Hang Zhang\textsuperscript{\rm 3}}, \Large \textbf{Gerard de Melo\textsuperscript{\rm 1}} \\ 
\textsuperscript{\rm 1}Rutgers University   
\textsuperscript{\rm 2} The Chinese University of Hong Kong 
\textsuperscript{\rm 3} Amazon AI
}
\hypersetup{draft}
\begin{document}

\maketitle

\begin{abstract}
Different from short videos and GIFs, video stories contain clear plots and lists of principal characters. Without identifying the connection between appearing people and character names, a model is not able to obtain a genuine understanding of the plots. Video Story Question Answering (VSQA) offers an effective way to benchmark higher-level comprehension abilities of a model. However, current VSQA methods merely extract generic visual features from a scene. With such an approach, they remain prone to learning just superficial correlations. In order to attain a genuine understanding of who did what to whom, we propose a novel model that continuously refines character-aware relations. This model specifically considers the characters in a video story, as well as the relations connecting different characters and objects. Based on these signals, our framework enables weakly-supervised face naming through multi-instance co-occurrence matching and supports high-level reasoning utilizing Transformer structures. We train and test our model on the six diverse TV shows in the TVQA dataset, which is by far the largest and only publicly available dataset for VSQA. We validate our proposed approach over TVQA dataset through extensive ablation study. 

\end{abstract}

\section{Introduction}

\begin{figure}[t!]
 \centering
 \includegraphics[width=1.0\columnwidth]{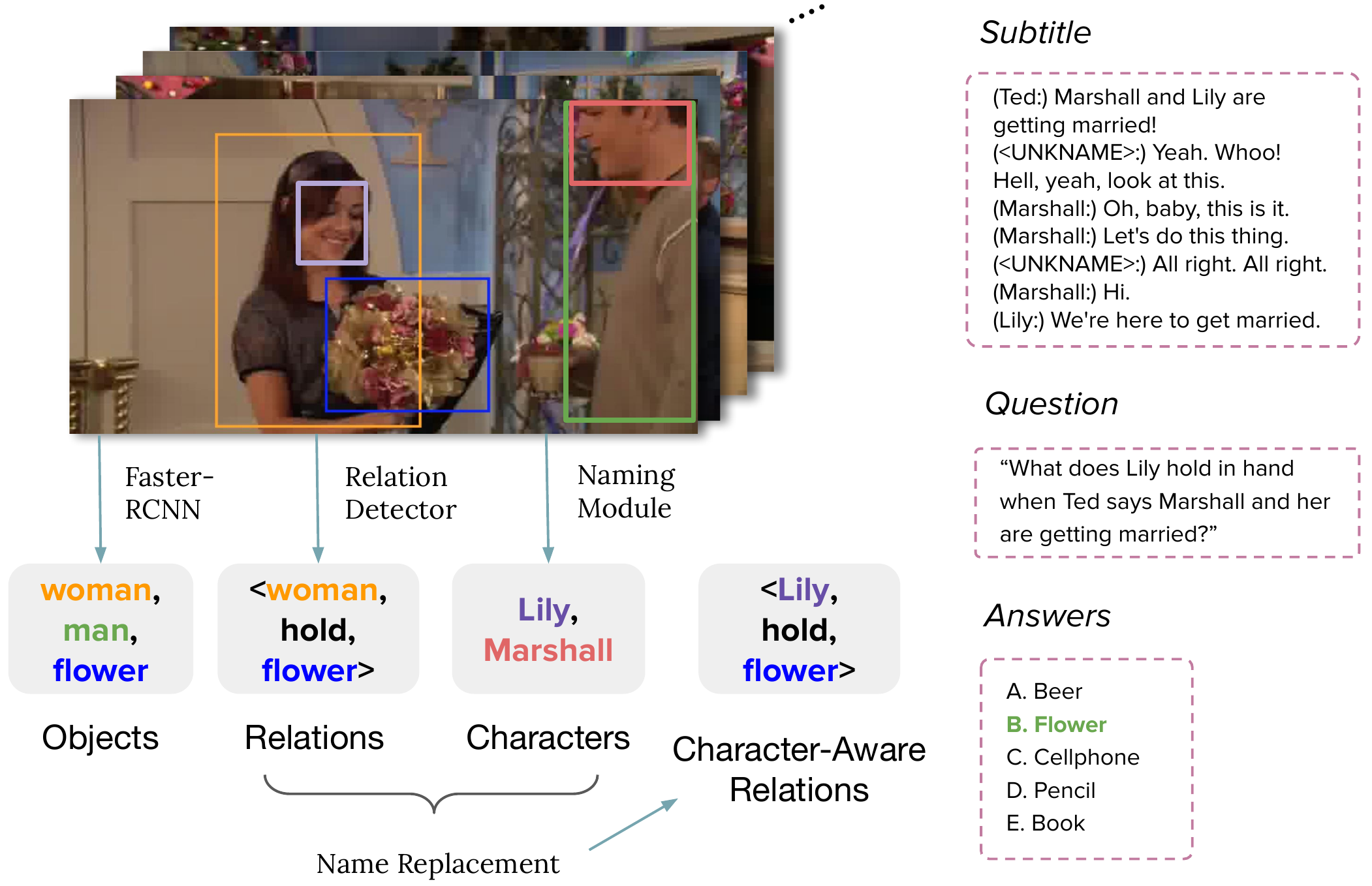}
 \captionsetup{font=small}
 \caption{Illustration of how fine-granular features may help in VSQA. Character names, visual objects, and their relationships are all necessary factors in answering this question. Character-aware relationships are detected in video frames, where references to humans such as ``woman'' and ``man'' are replaced with predicted character names, determined by finding the face bounding box that overlaps the most with the human bounding box.}
\label{fig:vrd_met}
\end{figure}


Video stories such as TV shows and movies entertain us and enrich our life. We can easily understand the plots and become addicted to the acting of protagonists. However, video story understanding remains a challenging task for artificial intelligence. In this paper, we argue that characters in two aspects play an important role for a better comprehension of video stories. On the one hand, characters lie at the intersection of video and text/subtitle modalities. On the other hand, they are the pivots of plots, embodying who did what to whom. 

The task of VSQA is a convincing means of measuring how well a model understands a video. Typically, it is solved in three steps: 1) extracting key features of multimodal contents; 2) fusing those multimodal features; 3) utilizing the fused features to predict the correct answer to a question. For the first step, current state-of-the-art methods \cite{Na_2017_ICCV,wang2018movie,kim2018multimodal,li2019beyond,kim2019gaining,kim2019progressive} mainly focus on global visual features at the image level. In particular, they consider one or more frames as input and extract features that provide a holistic representation of the frames. As a result, a basic understanding of what occurs in the frames is achieved, but substantially meaningful details may be missed due to the coarse granularity of the global features. Such details include individual objects, their relationships and attributes, and perhaps more importantly, the identities of people inside the video. These aspects are often crucial for answering semantic questions such as ``What does Lily hold in hand when Ted says Marshall and her are getting married?'' (Figure~\ref{fig:vrd_met}). Here, the flower (object) that Lily (character) is holding (relationship) are the key factors needed to answer the question, but generic global features usually have very limited power to capture them. Such limitations motivate us to design a framework that focuses on fine-grained visual cues and provides richer knowledge of the depicted scenes.

Several recent works~\cite{anderson2018bottom,lei2018tvqa,yao2018exploring} have followed this direction and explored various approaches to incorporate grounded visual features for questions answering tasks.
Despite their success, the video story setting is quite different in that characters' identities, especially those of main actors, tend to matter much more than in static images, since they keep reappearing. 
Moreover, video stories involve a much larger number of interactions between characters, such as ``Robin talks to Ted'', or between characters and objects, such as ``Lily holds flowers''. It is very difficult for a model to achieve a genuine understanding of a scene without capturing these character-involved associations. Therefore, we need a better framework that has the capability to mine both detailed visual cues about character identities and their relationships. 


To address the above issues, we build a VSQA framework accounting for character-centric relations and a character-aware reasoning network (CA-RN) that combines and connects those features with reasoning abilities. Our framework consists of two main parts. The first part aims at building a scene representation for understanding relations between characters and objects so as to infer what is going on. Through visual relations, we capture two levels of visual semantics: the entity level and the relation level. At the entity level, we detect characters, objects, and their attributes via pre-trained object detectors and multi-instance co-occurrence matching based character identification. At the relation level, the relations between the entities are recognized within each frame, where human-referring words are replaced with predicted character names. For the second part, the multi-modal information (including two-level scene representation and subtitles) are then injected into our Transformer-based CA-RN network, which serves as the semantic reasoning module. 

We train and test our model on the six diverse TV Shows from a large-scale video story dataset TVQA\footnote{\href{http://tvqa.cs.unc.edu/}{\url{http://tvqa.cs.unc.edu/}}. TVQA offers a large number (21.8K) of video clips from six TV shows -- Big Bang Theory (\emph{BBT}), \emph{Friends}, How I Met Your Mother (\emph{HIMYM}), Grey's Anatomy (\emph{Grey}), House M.D.\ (\emph{House}), and \emph{Castle}.}. In each video clip, there are corresponding subtitles and several multiple choice questions. The goal of our framework is to correctly predict the right answers to these questions. The key contributions of this paper can be summarized as follows:
\begin{itemize}
\itemsep0em
\item We propose an end-to-end multi-task framework to mining the face--speaker associations and conduct multi-modal reasoning at the same time. It enables weakly-supervised face naming through co-occurrence matching and supports high-level reasoning through Transformer structures.
\item We propose to utilize character-aware relations as a stronger representation of visual knowledge of scenes. To the best of our knowledge, this is the first attempt to apply such a strategy to video question answering tasks.
\item Experiments on six TV shows confirm that our approach outperforms several strong baselines while also offering explicit explanations, especially for those questions that require a deep understanding of video scenes.
\end{itemize}
\section{Related Work}

\textbf{Video Story Question Answering.} The task of video question answering has been explored in many recent studies. While some of them \cite{li2019beyond,gao2018motion,jang2017tgif} principally focus on factual understanding in short videos, another research direction aims at understanding videos that contain story-lines and answering questions about them. Read Write Memory Networks (RWMN)~\cite{Na_2017_ICCV} rely on Compact Bilinear Pooling to fuse individual captions with corresponding frames and store them in memory slots. Multi-layered CNNs are then employed to represent adjacent slots in time. 
PAMN~\cite{kim2019progressive} proposes a progressive attention memory to progressively prune out irrelevant temporal parts in memory and utilizes dynamic modality fusion to adaptively determine the contribution of each modality for answering questions.
ES-MTL~\cite{kim2019gaining} introduces additional temporal retrieval and modality alignment networks to predict the time when the question was generated and to find associations of video and subtitles.
However, these methods merely extract visual features from video frames or parts of video frames with pre-trained CNNs while ignoring the characters inside video scenes, making their models lack the ability of deep scene understanding.
\begin{figure*}[t!]
  \centering
  \includegraphics[width=0.83\textwidth]{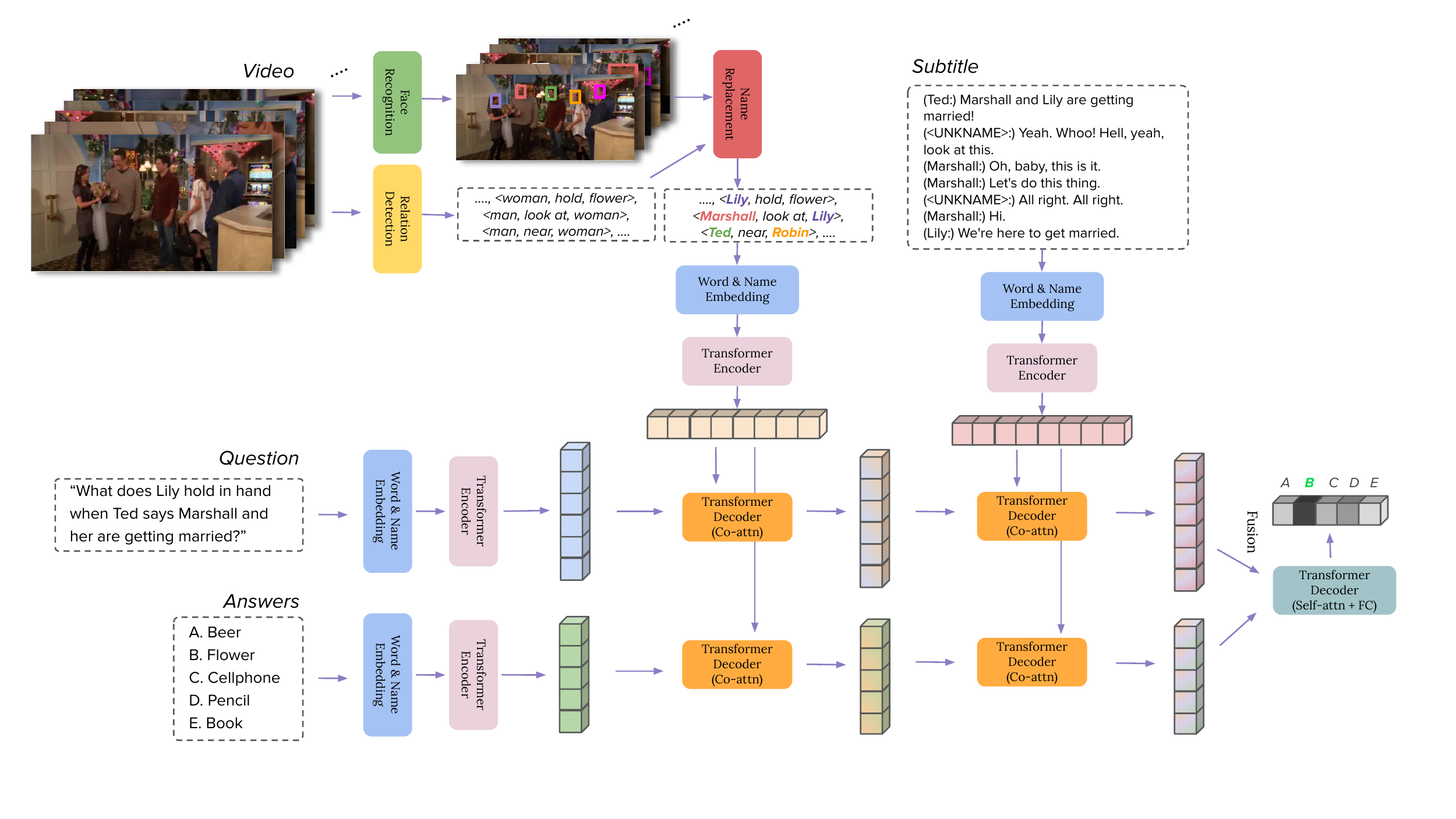}
\captionsetup{font=small}
\caption{Architecture of our CA-RN model. For each frames in a video, characters, objects, and visual relations are detected at first. Human-referring words in the visual relations are then replaced with predicted character names. The ordinary words and character names are embedded by two different embeddings. Finally, question, answer options, subtitles, and video semantics are fed into a transformer-based multi-modal reasoning structures to make answer predictions.}
\label{fig:model}
\end{figure*}

\noindent\textbf{Visual Relation Detection.} Visual relation detection has recently emerged as a task that goes one step further than object detection towards a holistic semantic understanding of images \cite{lu2016visual,krishna2017visual,xu2017scenegraph,openimages}. The task involves first detecting any visually related pairs of objects and recognizing the predicate that describes their relations.
Most recent approaches achieve this goal by learning classifiers that predict relations based on different types of features of the object pairs \cite{xu2017scenegraph,yu2017visual,zellers2018scenegraphs,zhang2018large,zhang2019vrd}. 
It has been demonstrated in recent works that scene graphs can provide rich knowledge of image semantics and help boost high-level tasks such as Image Captioning and Visual Question Answering \cite{yao2018exploring,liang2019rethinking,yang2018multi,gao2019dynamic,gao2018question,gao2019multi,geng20192nd,shi2020multi}. We are interested in how relations can be exploited not just for images but also for video understanding with a character-based relation representation, which to the best of our knowledge has not been fully explored yet.
 
\noindent \textbf{Character Naming.} The goal of character naming is to automatically identify characters in TV shows or movies. Previous methods tend to train a face assignment model based on extracted face tracklets. Some approaches rely on semi-supervised learning for person identification \cite{tapaswi2015improved,parkhi2018automated,bauml2013semi}. Meanwhile, \citet{jin2017end} propose an unsupervised method to address the task. In this work, we train the character naming and question answering modules in a multi-task scheme. Our approach does not require any explicit annotations on faces. We only rely on weak supervision from the subtitles that contain speakers' names and exploit the co-occurrence distribution between appearing faces and names in subtitles.

 

\section{The Proposed Method}

The goal of this work is to make use of the co-occurrence of faces in videos and names in subtitles to continuously refine the detection of character-aware relationships, and finally use the latter for improved video story understanding.
As shown in Figure~\ref{fig:model}, our video story understanding framework can be trained in an end-to-end manner and consists of two main modules, one of which predicts the detected face bounding boxes and incorporates character names into the detected relationships by matching the locations of bounding boxes. 
As a result, multiple forms of visual semantics from each frame are extracted and combined together as an understanding of the scene that the characters are acting in. The other module is a sequential Transformer-based reasoning pipeline, which takes in the input question, answer options, and different modalities, and outputs the predicted answer with the highest softmax score. In the following, we describe the methods to extract character-aware visual semantics and conduct multi-modal reasoning.


\begin{figure*}[t!]
  \centering
  \includegraphics[width=0.84\textwidth]{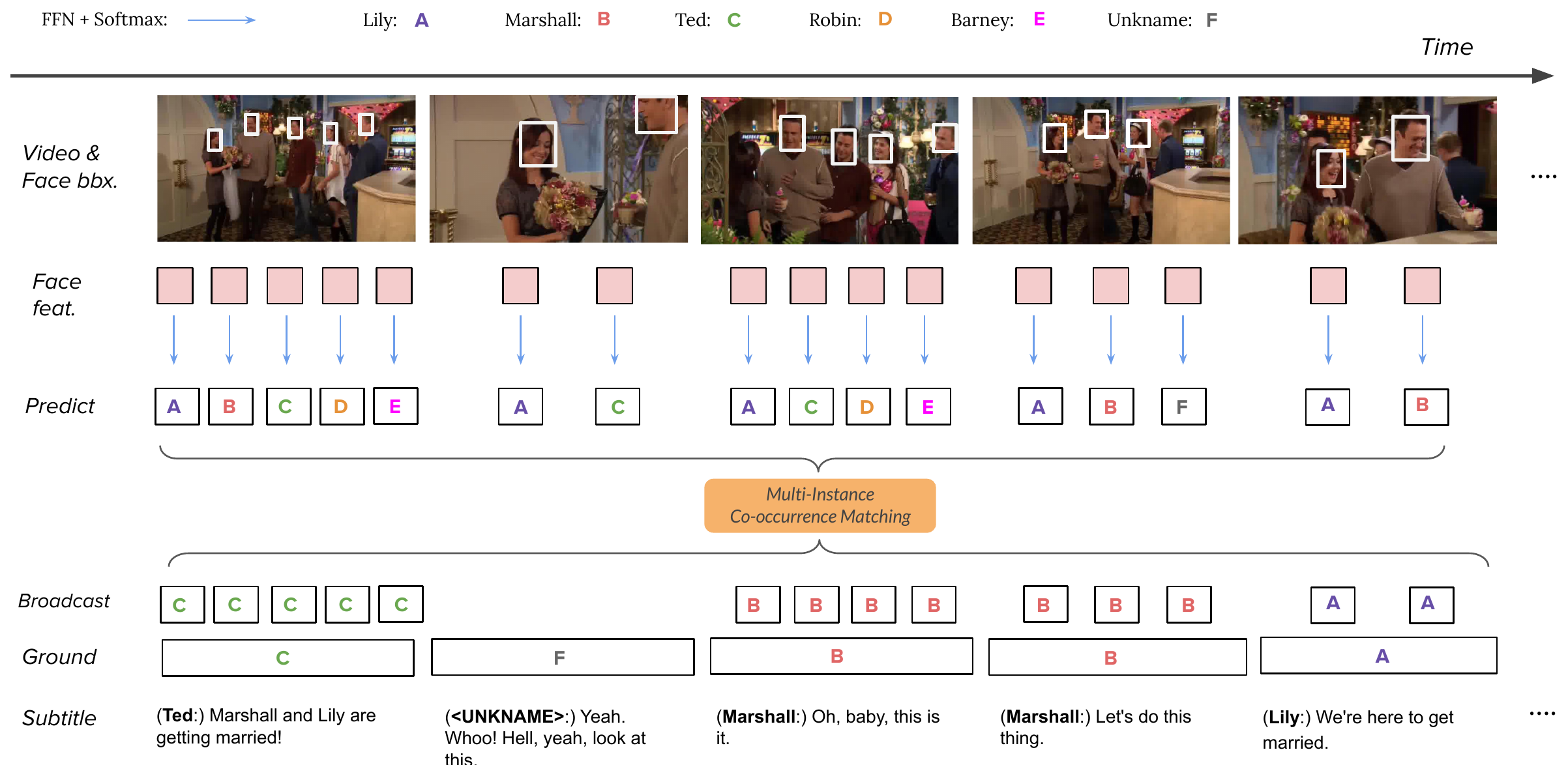}
\captionsetup{font=small}
\caption{An illustration of our weakly-supervised character identification pipeline. The face bounding boxes of all characters are first detected. The extracted face features are then predicted by fully-connected feed-forward layer and Softmax. After broadcasting the character names in subtitle to be a distribution sequence that has the same length of predicted name distribution sequence, a weight KL divergence loss is utilized to conduct multi-instance co-occurrence matching.
}
\label{fig:co-occurrence}
\end{figure*}

\subsection{Character-Aware Frame Understanding}
\label{sec:character_aware_frame_understanding}

\textbf{Face Detection and Feature Extraction.} We utilize a state-of-the-art face detector \cite{zhang2017s3fd} to localize faces in each frame and extract their 256-dimensional visual features $f \in \mathbb{R}^{256}$ using LightCNN \cite{wu2018light}, considering its effectiveness at general face identification, i.e., identifying different faces of the same person. This is a desirable feature, as we need to distinguish the faces of different people while neglecting the variance within the various appearances of a given person.

\noindent \textbf{Weakly Supervised Character Identification.} In order to recognize characters without explicit face name annotations, we first determine the number $k$ of principal characters in the TV series (details in Section \ref{sec:implement_details}). Supporting actors are lumped together as an UNKNAME class here, as they have a smaller impact on the main plot lines. 
Assume there are $n$ detected faces in a video clip with features $\mathcal{F}=\{f_1, ..., f_n\}$. We first utilize a naming module consisting of several fully connected feed-forward layers and softmax to get a confidence distribution over all names in the character list:
\begin{equation} 
\vspace*{-3pt}
    p_i = \mathrm{softmax}(W_2\mathrm{ReLU}(W_1f_i + b_1) + b_2))
\end{equation}%
\vspace*{-2pt}%
\noindent where $W_1$, $W_2$, $b_1$, $b_2$ are the weights and biases of fully-connected layers, and $p_i$ is the confidence distribution over all names in the character list. By doing so for all detected faces, we can construct a sequence of predicted character name distributions: $\mathcal{P}=\{p_1,p_2,..., p_n\}$.
As shown in Figure~\ref{fig:co-occurrence}, the speaker names in the subtitle maintain a multi-instance co-occurrence with the character faces in the video. Inspired by this, we duplicate the current speaker name to be of the same number as the detected faces in each frame to serve as weak supervision. Note that if the speaker in the subtitle is UNKNAME, the frame will not have a broadcast operation. The ground character name distribution can be represented by $\mathcal{G}=\{g_{\mathrm{Loc}(1)},g_{\mathrm{Loc}(2)},..., g_{\mathrm{Loc}(n)}\}$, 
where the localization function $\mathrm{Loc()}$ maps the face bounding box ID to the frame ID, and $g_l$ denotes the one-hot ground name distribution of frame $l$.
Afterwards, the multi-instance co-occurrence matching can be conducted by a regularized Kullback-Leibler divergence between predicted and ground character name distributions:

\begin{equation} 
D_{\mathrm{RKL}}(\mathcal{P}\parallel \mathcal{G}) = \sum_{l}^L \min_{j \in \mathrm{F}_l} \mathcal{P}(j)\ln\frac{\mathcal{P}(j)}{\mathcal{G}(j)} \label{eq:char2}
\end{equation}
\noindent where $\mathrm{F}_l$ is the set of faces in frame $l$. This loss is similar in spirit to Multiple Instance Learning. With this, the model learns to assign the speaker name to the corresponding face, which will minimize the loss.

\noindent \textbf{Regular Visual Object Contexts.} We use regular objects and attributes as another form of visual semantics for each frame, similar to \citet{lei2018tvqa}. Specifically, we apply Faster-RCNN \cite{ren2015faster} trained on Visual Genome \cite{krishna2017visual} to detect all objects and their attributes. Object bounding boxes are discarded, as we are targeting pertinent semantic information from the frame.

\noindent \textbf{Incorporating Characters into Visual Semantics.} Once we have localized and recognized character faces and names, as well as regular objects, we append each character name detected in a given frame to each of the objects detected in the same frame to augment its visual semantics. We found that this simple strategy works very well as shown in Section~\ref{sec:abl}, due to the fact that character names are anchors that allow for localizing and disentangling semantic information. For example, visual objects without names attached, such as ``food, wine glass'', are fairly generic, while ``food+Leonard'' and ``wine glass+Penny'' better allow for distinguishing objects from different frames and associating them with relevant people, and hence provide a clearer picture of what is included in the scene. 


\subsection{Character-Aware Relation Detection}
This component is designed to extract all relationships that involve the main characters in the scene. We decompose this task into two steps: 1) detecting relationships to build scene graphs; 2) replacing detections of humans in  relationships with specific characters. 

\noindent \textbf{General Relation Detection.} The relation detection module aims at detecting all related objects and recognizing their relationships in a video. We use the approach by \citet{zhang2018large} to detect relationships in each frame. Specifically, we train the model on the VG200 dataset, which contains 150 objects and 50 predicates, and apply this model on all frames of a given video. The result is a set of $\langle S,P,O \rangle$ triples per frame, where $S$, $P$, $O$ represent the \textit{subject}, \textit{predicate}, \textit{object}, respectively. We only keep the $\langle S,P,O \rangle$ triples and discard subject/object boxes since 1) their spatial locations carry little signal about the scene semantics; 2) there are already many spatial relationships among the VG200 predicates, such as ``above'' and ``under''. Once the model manages to predict these relations, we immediately know the relative spatial relations between subjects and objects, which we found is sufficient to describe the scenes. We concatenate all the triples in the current frame into a sequence of $N_r \times 3$ (where $N_r$ is the number of $\langle S,P,O \rangle$ triples) and feed it to the following modules.

\noindent \textbf{Character Name Replacement.} Given all the faces and relations in a scene, we focus on those relations with human-referring words as subjects or objects, such as ``woman'' or ``man''. For each of these human bounding boxes, we obtain the face box that overlaps the most with it to determine the human's face. Once this matching is done for all human boxes in the frame, we replace the human-referring words in those relationships with the previously identified character names. This makes the relationships more concrete, as we know exactly who is involved in each relationship. Figure \ref{fig:vrd_met} shows examples of detected relationships from frames of \textit{HIMYM}, where human-referring words are replaced with the specific character names. We show in Section~\ref{sec:abl} that by applying this name replacement to all relationships, the model is able to capture details that are strongly associated with the question and thus engender more accurate answers.



\subsection{Character-Aware Reasoning Network}
As shown in Figure~\ref{fig:model}, CA-RN model works in an end-to-end manner and updating based on a multi-task loss function. It takes in question, answer options, subtitles, face bounding boxes, and visual semantics, and then outputs the probabilities of each answer option. Motivated by the Intra-Inter Attention~\cite{gao2019dynamic,gao2019multi} flow mechanism, we propose a transformer-based multi-modality fusion apparoch for Character-Aware Reasoning Network.

\noindent \textbf{Encoder.} We first embed the input question, answer options, and all modalities (subtitles and visual semantics) $\mathcal{I} = \{q, a_{0-4}, s, v_{o,r}\}$ using word and name embeddings. We denote the set of these embeddings as: $\mathcal{E} = \{e_q, e_a, e_s, e_v\}$. They are then fed into a two-layer Transformer encoder consisting of self-attention with four heads to capture long-range dependencies and a bottle-neck feed-forward networks to obtain encoded hidden representations:
\begin{equation}
    h_j = \mathrm{FFN}(\mathrm{Attention}(e_j)),
\end{equation}
where $j \in \{q, a, s, v \}$, FFN is a feed-forward module consisting of two FC layers with ReLU in between, and the Attention function here is defined as~\cite{vaswani2017attention}:
\begin{equation}
    \mathrm{Attention}(Q,K) = \mathrm{softmax}(\frac{QK^T}{\sqrt{d_h}})K.
\end{equation}
Here $\sqrt{d_h}$ is a scaling factor used to maintain scalars in the order of magnitude and $d_h$ is each head's hidden dimensionality.

\noindent \textbf{Multi-Modal Decoder.} Once all inputs are encoded by the encoder, we utilize sequential co-attention decoders to fuse their information and generate an updated representation of the question and answer options. For simplicity, we take visual relations $h_r$ and subtitles $h_s$ as two input modalities. The following framework can easily be extended to more input modalities. As shown in Figure \ref{fig:model}, the visual relations, question, and answer options are first fed into the a two-layer four-heads co-attention decoder to acquire context-aware-QA representations. Then, subtitles and updated QA representations serve as the input for another co-attention decoder with the same structure. The co-attention decoder can be represented as:
\begin{equation}
    h_{c\rightarrow i} = \mathrm{FFN}(\mathrm{Attention}(h_i, h_c)),
\end{equation}
where $h_{c\rightarrow i}$ is the context-aware-QA representation, and $h_c$, $h_i$ represent contextual and input QA hidden representations, respectively.
Afterwards, context-aware QA representations for different question--answer pairs are then concatenated and processed by a self-attention decoder to get the final representation for softmax calculation:
\begin{equation}
    \begin{aligned}
    	M & ~=~ \mathrm{Concat}_{i \in \{(q,a_0),...,(q,a_4) \}} h_{c\rightarrow i}\\
    	p_a & ~=~ \mathrm{softmax}(\mathrm{FFN}(\mathrm{Attention}(M)))\\
    \end{aligned}
\end{equation}
Finally, we are able to predict the answer $y$ with the highest confidence score $y = \argmax_{a \in \{a_0,...,a_4\}} p_a$.

\noindent \textbf{Multi-Task Loss Function.} The feed-forward network in the naming module and the multi-modal reasoning module are jointly trained in an end-to-end manner through the following multi-task loss function:
\begin{equation}
   \begin{aligned}
    \mathcal{L}_\mathrm{multi-task} & ~=~ \mathcal{L}_\mathrm{cross-entropy} + \lambda \mathcal{L}_\mathrm{mi-co} \\
    & ~=~ -\sum_{c=1}^{5} g_c \log(p_c) + \lambda D_{\mathrm{RKL}}(\mathcal{P}\parallel \mathcal{G}),
   \end{aligned}
\end{equation}
which combines $\mathcal{L}_\mathrm{cross-entropy}$ for question answering and $\mathcal{L}_\mathrm{mi-co}$ for multi-instance co-occurrence matching, linked together by a hyperparameter $\lambda$.

\begin{table*}[t!]
\centering
\resizebox{0.8\textwidth}{!}{
\begin{tabular}{l|cccccc|c|c}
\hline
\textbf{w/ ts} & \multicolumn{7}{c|}{\textbf{Test-Public}} & \textbf{Val} \\
\hline 
Show & BBT  & Friends & HIMYM & Grey & House & Castle & All & All \\
\hline
NNS-SkipThought \cite{lei2018tvqa} & - & - & - & - & - & - & 38.29 & 38.41 \\
NNS-TFIDF \cite{lei2018tvqa} & - & - & - & - & - & - & 50.79 & 51.62 \\
Multi-Stream V only \cite{lei2018tvqa} & - & - & - & - & - & - & 43.69 & - \\
Multi-Stream \cite{lei2018tvqa} & 70.19 & 65.62 & 64.81 & 68.21 & 69.70 & 69.79 & 68.48 & 68.85 \\
T-Conv~\cite{yu2018qanet} & 71.36 & 66.52 & \textbf{68.58} & 69.22 & 67.77 & 68.65 & 68.58 & 68.47 \\
\hline
CA-RN (Ours) & \textbf{71.89} & \textbf{68.02} & 67.99 & \textbf{72.03} & \textbf{71.83} & \textbf{71.27} & \textbf{70.59} & \textbf{70.37} \\
\hline
Human & - & - & - & - & - & - & 91.95 & 93.44 \\
\hline
\end{tabular}
}
\captionsetup{font=small}
\caption{Results on the TVQA test setor models that use time-stamp annotation (`w/ ts'). We compare to other baselines on the six TVQA sub-datasets individually. ``V only'' means using only global CNN features without subtitles.
}
\label{tab:with}
\end{table*}
\begin{table*}[t!]
\centering
\resizebox{0.8\textwidth}{!}{
\begin{tabular}{l|cccccc|c|c}
\hline
\textbf{w/o ts} & \multicolumn{7}{c|}{\textbf{Test-Public}} & \textbf{Val} \\
\hline 
Show & BBT  & Friends & HIMYM   & Grey  & House  & Castle & All & All \\
\hline   
NNS-SkipThought \cite{lei2018tvqa} & - & - & - & - & - & - & 26.93 & 27.50 \\
NNS-TFIDF \cite{lei2018tvqa} & - & - & - & - & - & - & 49.59 & 50.33 \\
Multi-Stream V only \cite{lei2018tvqa} & - & - & - & - & - & - & 42.67 & - \\
Multi-Stream \cite{lei2018tvqa}  &  70.25 &	65.78 &	64.02 &	67.20 &	66.84 &	63.96 &	66.46 & 65.85 \\
T-Conv~\cite{yu2018qanet} & 67.38 & 63.97 & 62.17 & 65.19 & 65.38 & 67.88 & 65.87 & 65.85 \\
PAMN~\cite{kim2019progressive} & 67.65 & 63.59 &	62.17 &	67.61 &	64.19 &	63.14 &	64.61 & 64.62 \\
ES-MTL~\cite{kim2019gaining}  & 69.60 &	\textbf{65.94} &	64.55 &	68.21 &	66.51 &	66.68 &	67.05 & 66.22 \\

\hline
CA-RN (Ours)  & \textbf{71.43} & 65.78 & \textbf{67.20} & \textbf{70.62} & \textbf{69.10} & \textbf{69.14} & \textbf{68.77} & \textbf{68.90} \\
\hline
Human & - & - & - & - & - & - & 89.41 & 89.61 \\
\hline
\end{tabular}
}
\captionsetup{font=small}
\caption{Results on the TVQA test set for models that do not use time-stamp annotations (`w/o ts'). We compare to other baselines on the six TVQA sub-datasets individually. ``V only'' means using only global CNN features without subtitles.
}
\label{tab:without}
\end{table*}

\subsection{Implementation Details}
\label{sec:implement_details}



\noindent \textbf{Principal Character List.} We focus on naming the faces of principal characters, since they are highly correlated to the story-line of TV shows. The number $k$ for the character list for each TV show is determined in the following three steps: 1) count the occurrences of all speakers in the subtitles; 2) select all names appearing more than 500 times as principal character candidates; 3) filter out names that make up less than 1/10 of the speakers with the highest occurrence. 4) An additional UNKNAME class is assigned to all other character names not in the principal list. Note that we do not rely on external information of who the principal characters are.

\noindent \textbf{Text and Name Embeddings.} After parsing the scene into character-aware relations, we have four types of features to transform into text embeddings: the subtitles, visual semantics, questions, and the candidate answers. Note that once the visual semantics are extracted, we do not need any visual features from the frames. Since the character names are different from their literal meanings (e.g., the character Lily is different from the regular word \emph{lily}), we utilize two separate embeddings. For ordinary words, we rely on 300-dimensional GloVe word vectors \cite{pennington2014glove} to embed the words after tokenization. For character names, we train and update their name embeddings from scratch. In the case of out of vocabulary words, we use averaged character vectors of the words.

\noindent \textbf{Model Training.} Our model is trained with Adam stochastic optimization on Tesla V100 GPUs. The $\lambda$ in the multi-task loss function is simply set to 1. In the training process, we set the batch size as 64. The learning rate and optimizer settings are borrowed from \cite{lei2018tvqa}.

\section{Experiments}
\subsection{Dataset}
The recently released TVQA dataset~\cite{lei2018tvqa} is a large-scale video question answering dataset based on 6 popular TV shows: 3 situation comedies (The Big Bang Theory, Friends, How I Met Your Mother), 2 medical comedies (Grey's Anatomy, House M.D.), and 1 crime comedy (Castle). It consists of 152.5K QA pairs (84.8K what, 17.7K who, 17.8K where, 15.8K why, 13.6K how questions) from 21.8K video clips, spanning over 460 hours of video. Each video clip is associated with 7 questions and a dialogue text (consisting of character names and subtitles). The questions in the TVQA dataset are designed to be compositional in the format ``[What/How/Where/Why/...] \underline{\hbox to 7mm{}} [when/before/after] \underline{\hbox to 7mm{}}'' and require both visual and language comprehension.

\subsection{Baselines}
We consider several baselines for performance comparison.


\noindent \textbf{Nearest Neighbor Search.} These baselines (NNS-TFIDF and NNS-SkipThought) are taken from the original TVQA paper \cite{lei2018tvqa}. They compute the cosine similarity between the resulting vectors to predict the answer.

\noindent \textbf{Multi-Stream.}~\cite{lei2018tvqa} combines information from different modalities with LSTMs and cross-attention. The results stem from the official TVQA leaderboard.

\noindent \textbf{Temporal Convolution.} It has recently been shown that temporal convolutions (T-Conv) \cite{Na_2017_ICCV,yu2018qanet} can be a strong alternative to traditional RNN layers for question answering. We follow the structure from \cite{yu2018qanet} and build the T-Conv baseline by replacing the LSTM layers in \citet{lei2018tvqa} with temporal convolutions while keeping other modules unchanged.

\noindent \textbf{PAMN.} ~\citet{kim2019progressive} utilize progressive attention memory to update the belief for each answer. This is also from the official TVQA leaderboard.

\noindent \textbf{ES-MTL.} ~\citet{kim2019gaining} explores two forms of extra supervision for temporal localization and modality alignment. This is the strongest baseline from the official TVQA leaderboard without any additional object-level annotations.




\noindent \textbf{Human Performance.}
We also give the human results as reported along with the dataset \cite{lei2018tvqa} as a reference to gauge how big the gap is to human intelligence.


\begin{table}[t!]
\centering
\resizebox{0.8\columnwidth}{!}{
\begin{tabular}{lcc}
\hline\noalign{\smallskip}
 & \multicolumn{2}{c}{Val Acc.} \\
Method & w/ ts & w/o ts \\ 
\noalign{\smallskip}
\hline
\noalign{\smallskip}
Sub & 66.23 & 66.14 \\
Sub + Objs & 68.85 & 67.35 \\
Sub + Rels & 67.55 & 67.16 \\ 
Sub + Objs\_nm & 69.45 & 68.13 \\
Sub + Rels\_nm & 68.25 & 67.85 \\ 
Sub + Objs + Rels & 69.54 & 68.44\\ 
Sub + Objs + Rels\_nm & 69.76 & 68.64 \\ 
Sub + Objs\_nm + Rels & 70.20 & 68.68 \\ 
Sub + Objs\_nm + Rels\_nm & \textbf{70.37} & \textbf{68.90} \\
\noalign{\smallskip}
\hline
\end{tabular}
}
\setlength\belowcaptionskip{-3ex}
\captionsetup{font=small}
\caption{Ablation study both with and without the time stamps. ``Sub'', ``Objs'', ``Rels'' and ``nm'' represent subtitles, objects, relationships and names, respectively.}
\label{tab:ablation}
\end{table}


\begin{figure*}[t!]
  \centering
  \begin{subfigure}{\textwidth}
    \centering
    \begin{subfigure}{0.47\textwidth}
      \centering
      \includegraphics[width=0.82\textwidth]{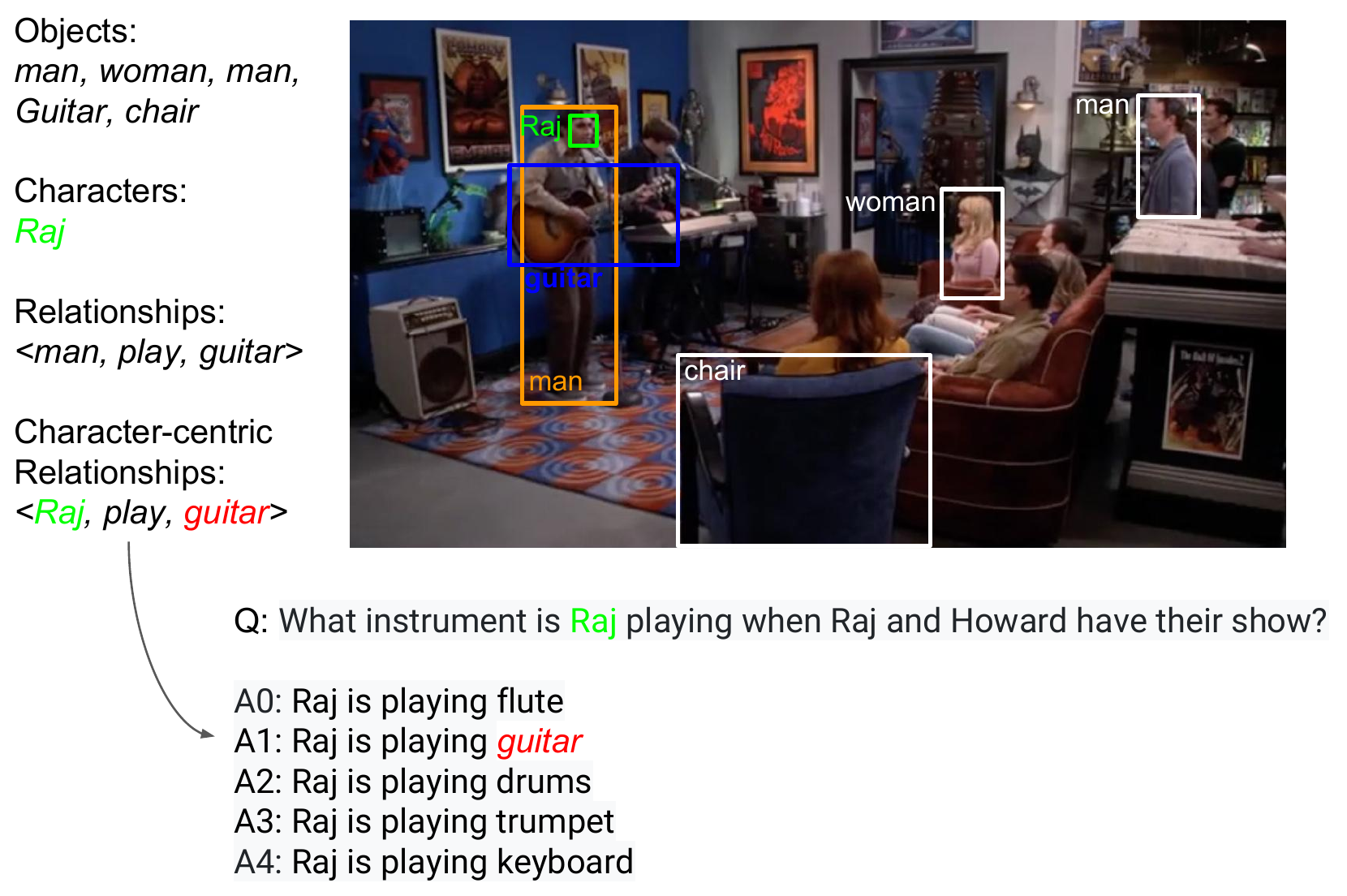}
    \end{subfigure}
    \centering
    \begin{subfigure}{0.47\textwidth}
      \centering
      \includegraphics[width=0.82\textwidth]{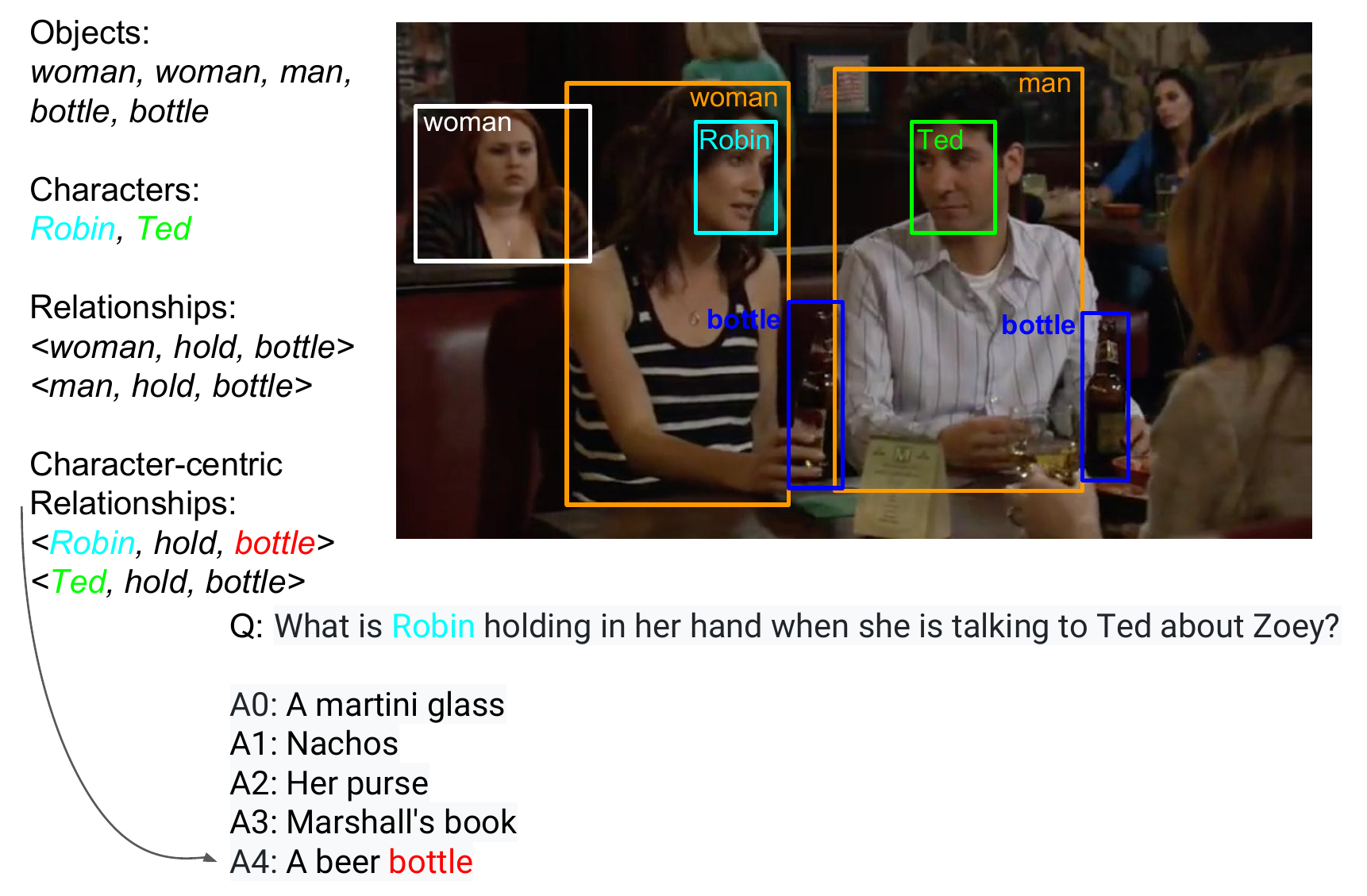}
    \end{subfigure}
  \end{subfigure}
  
  \centering
  \begin{subfigure}{\textwidth}
    \centering
    \begin{subfigure}{0.47\textwidth}
      \centering
      \includegraphics[width=0.82\textwidth]{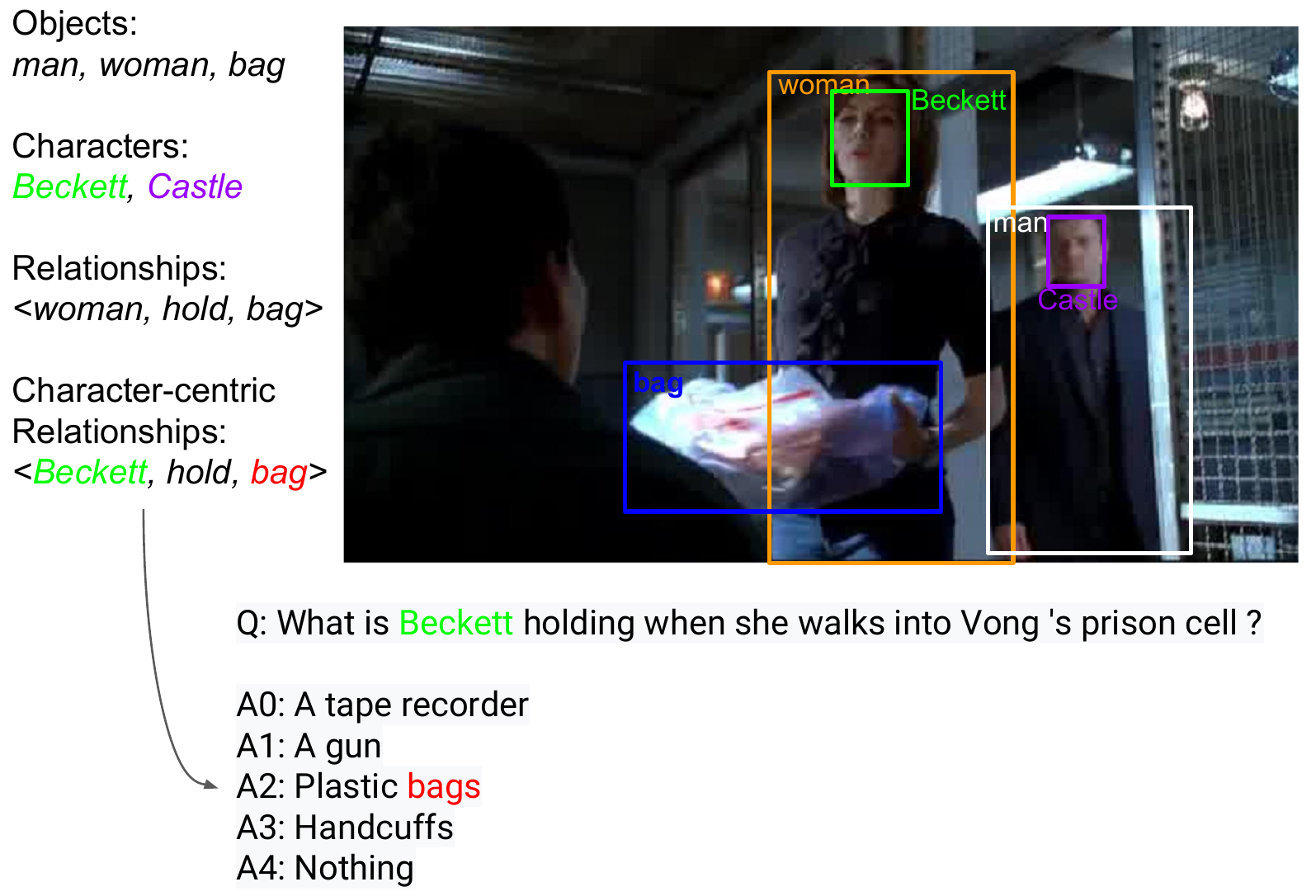}
    \end{subfigure}
    \centering
    \begin{subfigure}{0.47\textwidth}
      \centering
      \includegraphics[width=0.82\textwidth]{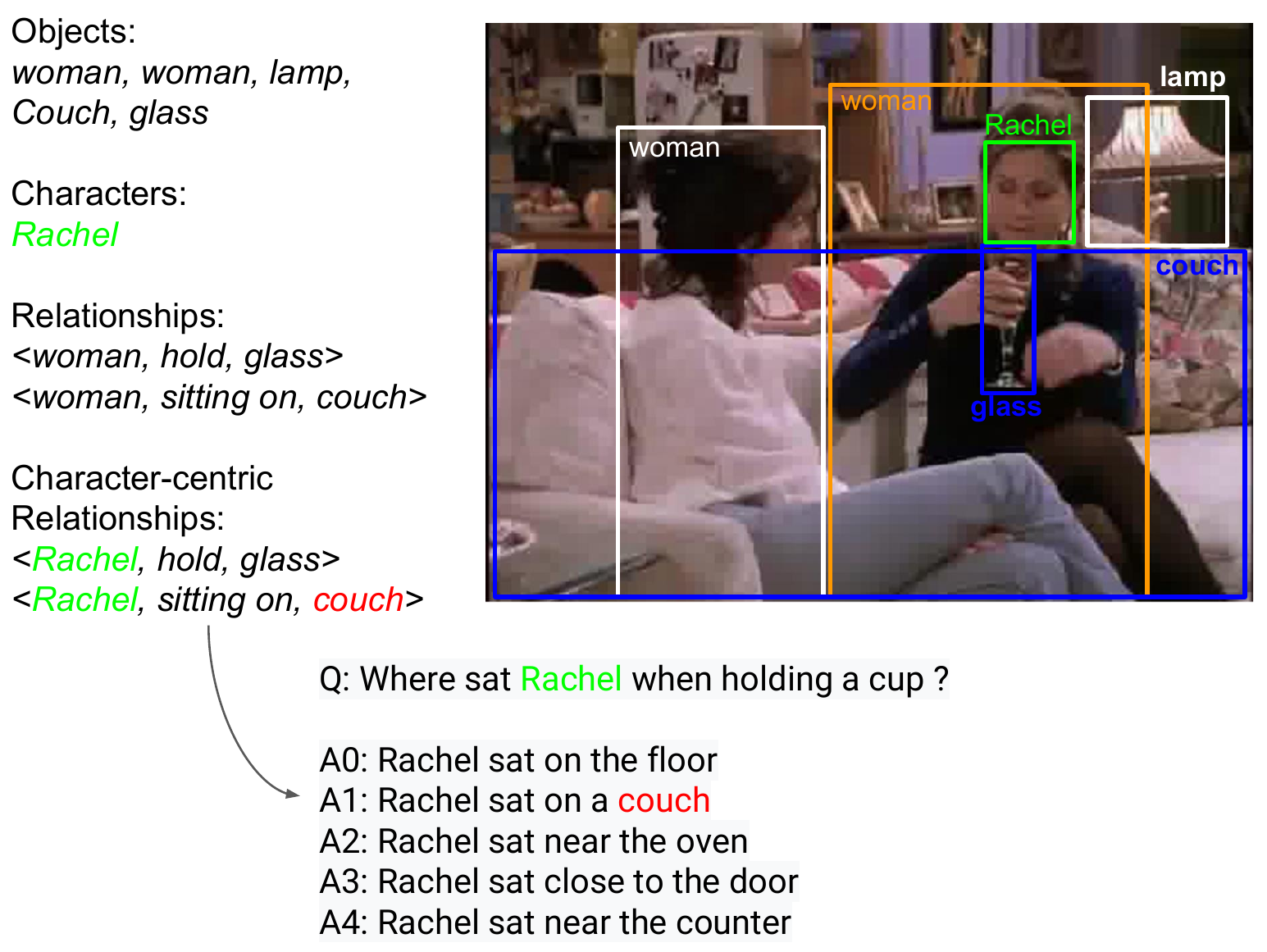}
    \end{subfigure}
  \end{subfigure}
\setlength\belowcaptionskip{-2ex}
\captionsetup{font=small}
\caption{Examples of correctly answered questions that benefit from the proposed strategy. Orange and blue boxes are subjects and objects, while white boxes are objects with no detected relationships. Boxes with names are our detected characters, which substitute for the human-referring words in the relationships to obtain a character-aware understanding.}
\label{fig:qualitative}
\end{figure*}

\subsection{Experimental Setup}

We use the top-1 accuracy as the only metric, following the official guidelines.
There are two types of settings we can adopt from the official evaluation rules \cite{lei2018tvqa}: with time stamps (w/ ts) and without time stamps (w/o ts), where time stamps refer to ground-truth annotations on the intervals of the video segments that relate the most to the given questions. The former setting assumes we have the time stamps in both training and testing, while in the latter case such information is not provided. 
We consider both settings in our comparison with related work.


\subsection{Comparison to State-of-the-art Approaches}
As presented in Tables \ref{tab:with} and \ref{tab:without}, our approach outperforms the best previous method by $1.90/2.01\%$ (absolute) on the val/test set with time stamps, and by $2.68/1.72\%$ (absolute) on the val/test set without time stamps. Considering that there are 15,253 and 7,623 validation and test questions, respectively, the largest gains are $15,253 \times 2.68\% = 409$ and $7,623 \times 2.01\% = 153$ questions on the two sets, respectively.
This establishes the strength of our multi-task character-aware reasoning network. We believe the reasons behind the performance boost are the following three: 1) the Transformer structure enables capturing longer dependencies within and between different modalities compared with traditional RNN structures, especially when there is a long subtitle. 2) The multi-task framework allows refining the multi-modal reasoning model and mining the correlation between faces and names at the same time, making the two tasks contribute to each other. 3) The character-aware relations offer more detailed information than global CNN features and enable a deeper scene understanding.

\subsection{Ablation Study}
\label{sec:abl}

For further analysis, we conduct an ablation analysis on our proposed model both with and without time stamps in Table~\ref{tab:ablation}. There are three forms of visual semantics that we incrementally combine together: objects (Objs), relationships (Rels), and character names (nm).
We observe that using subtitles only gives a reasonably good result of 66.23\% but is still significantly worse than approaches with visual semantics. When objects are added (``Sub+Objs''), the accuracy is boosted by 2.62\% (absolute), while additional gains (0.6\% absolute) occur with names added (``Sub+Objs\_nm''). A further improvement (0.92\%) is attained when character-centric relationships (``Sub+Objs\_nm+Rels\_nm'') are integrated, which demonstrates our claim that character naming is a beneficial factor for better video story understanding. We also provide results for further settings where we remove one or two forms of visual semantics, as in ``Sub+Rels'' and ``Sub+Rels\_nm''. These results are slightly worse than the counterparts using ``Objs'' instead of ``Rels'', for which we find two main causes: 1) Objects are usually more diverse than relations, since typically only a small subset of objects are related. 2)
Object detectors are generally more accurate than relationship detectors, which makes the ``Objs'' semantics more reliable than ``Rels''.

\subsection{Qualitative Results}
In Figure~\ref{fig:qualitative}, we present 4 examples of our model's results based on all forms of visual semantics.
In the top right case, the question demands a deep understanding of the scene where Robin is sitting beside Ted and holding a beer bottle.
The character-aware relations are particularly helpful when the question includes multiple relations, as in the bottom right example, where Rachel is sitting on a couch and holding a glass at the same time, requiring  models to learn this combination of relations in order to answer the question.
\section{Conclusion}
In this paper, we propose character-aware scene understanding for improved VSQA. Our character-aware reasoning network is trained in a end-to-end multi-task style to acquire weakly supervised character identification as well as video story understanding. For the experiments on the six TV shows, our full Subtitle + Objects + Relations + Names model achieves the best accuracy against all baselines, which confirms the effectiveness of our multi-task framework and character-aware reasoning model.
\bibliographystyle{aaai}
\bibliography{paper}

\end{document}